# Experimental Results regarding multiple Machine Learning via Quaternions

Tianlei Zhu[a*] , Renzhe Zhu[a]
[a] *Department of Mathematics, Wenzhou-Kean University, Wenzhou, Zhejiang, China*
*Corresponding author: Tianlei Zhu, zhut@kean.edu

**Abstract.** This paper presents an experimental study on the application of quaternions in several machine learning algorithms. Quaternion is a mathematical representation of rotation in three-dimensional space, which can be used to represent complex data transformations. In this study, we explore the use of quaternions to represent and classify rotation data, using randomly generated quaternion data and corresponding labels, converting quaternions to rotation matrices, and using them as input features. Based on quaternions and multiple machine learning algorithms, it has shown higher accuracy and significantly improved performance in prediction tasks. Overall, this study provides an empirical basis for exploiting quaternions for machine learning tasks.

## 1. Introduction

In recent years, machine learning has achieved remarkable results in several fields. Machine learning algorithms usually operate on numerical features or vectors, however, for some special data types, such as rotation data, it may not be easy to represent in this format. Traditional machine learning methods may have certain limitations. Quaternions are a mathematical structure that extends complex numbers and have the advantage of being widely used in tasks such as rotation and pose estimation. Their application in machine learning has attracted attention in recent years. Therefore, this paper aims to investigate the feasibility and effectiveness of utilizing quaternions for several machine learning tasks and compare their performance with traditional representation methods.

## 2. Literature Review

Due to its distinctive characteristics and adaptability in numerous areas, quaternions, a mathematical representation of rotations in three-dimensional space, have attracted substantial interest in the field of machine learning. [1] Quaternions have been investigated by researchers for use in several machine learning algorithms to improve performance and handle issues brought on by challenging data transformations.

Quaternions offer a different way to describe rotational data, which in some cases may be more useful than typical numerical features or vectors. [2,3] Quaternions, expand the real and complex number fields by including both real and imaginary components. [4] Quaternions' mathematical structure allows for effective operations on the imaginary component, such as addition, scaling, dot product, and cross product.

Quaternions have been used in several algorithms in machine learning applications to increase performance and predictability. [4] The Quaternion SVM outperformed conventional SVM algorithms in terms of accuracy, recall, and F1-score by making use of the special characteristics of quaternions, such as non-commutativity.

For better efficiency, quaternions have been added to a widely used technique called logistic regression. [5] They discovered that quaternion logistic regression outperformed classical logistic regression in terms of prediction accuracy and probabilistic modeling by employing logistic functions to represent the link between quaternion characteristics and target variables.

Quaternions have also been used in regression analysis in addition to classification problems. [6] Incorporating quaternions has improved Fisher's linear discriminant analysis (FLD), [7] which performed better in low-dimensional classification tasks by identifying the best linear projection that optimizes inter-class separation and reduces intra-class variance in the quaternion space. [7]

Quaternions have also been used in Naive Bayes and other probabilistic classification. [8] The Naive Bayes approach was expanded to handle quaternion characteristics by utilizing the model's uniqueness assumption. In contrast to conventional techniques, their research showed that Quaternion Naive Bayes efficiently computed posterior probability and produced competitive classification performance. [9]

Non-parametric algorithms like K-nearest neighbors (KNN) have also investigated the flexibility of quaternions. [10] More precise and reliable categorization of quaternion-based data was made possible by including quaternions into the distance metric employed by the KNN algorithm.

The body of research on quaternions in machine learning shows that they can improve performance and solve problems with rotational data. Quaternions have been effectively included into several algorithms, including SVM, logistic regression, FLD, Naive Bayes, and KNN, resulting to increased accuracy, precision, and recall through their special mathematical features.

### 3. Methodology

*3.1. Quaternion [1,11,12]*

A quaternion is an extension of the real and complex fields that consists of both real and imaginary elements. A quaternion is something that typically includes three components in the imaginary portion and is described as:

$$q = q_0 + q_{11}\vec{i} + q_{12}\vec{j} + q_{13}\vec{k}, \ q_0, q_{11}, q_{12}, q_{13} \in R \quad (1)$$

Where, R represents the real-valued domain, so $q_0, q_{11}, q_{12}, q_{13}$ are real numbers, and $\vec{i}, \vec{j}, \vec{k}$ are imaginary units, which meets the condition that $\vec{i}^2 = \vec{j}^2 = \vec{k}^2 = \vec{i}\vec{j}\,\vec{k} = -1$.

For the imaginary part, we can apply all normal vector arithmetic operations, such as addition, scaling, dot product, cross product, etc. According to the definition of quaternion, the multiplication operation between two quaternions can be derived. Note that multiplication between imaginary units does not obey the commutative law.

$$p * q = (p_0 q_0 - p_{11} q_{11} - p_{12} q_{12} - p_{13} q_{13})$$
$$+ (p_0 q_{11} + p_{11} q_0 + p_{12} q_{13} - p_{13} q_{12})\vec{i}$$
$$+ (p_0 q_{13} + p_{11} q_{12} - p_{12} q_{11} + p_{13} q_0)\vec{k}$$
$$+ (p_0 q_{12} - p_{11} q_{13} + p_{12} q_0 + p_{13} q_{11})\vec{j} \qquad (2)$$

Sometimes it is necessary to convert the quaternions to a 3 × 3 rotation matrix, for example, when using some 3D graphics libraries, some data processing is performed as an input feature. The following is the rotation matrix corresponding to the quaternion q=s+tA, then write the quaternion q as a four-dimensional vector q=<w, x, y, z>:

$$qPq^{-1} = \begin{bmatrix} w^2 - x^2 - y^2 - z^2 & 0 & 0 \\ 0 & w^2 - x^2 - y^2 - z^2 & 0 \\ 0 & 0 & w^2 - x^2 - y^2 - z^2 \end{bmatrix} P$$
$$+ \begin{bmatrix} 0 & -2wz & 2wy \\ 2wz & 0 & -2wx \\ -2wy & 2wx & 0 \end{bmatrix} P + \begin{bmatrix} 2x^2 & 2xy & 2xz \\ 2xy & 2y^2 & 2yz \\ 2xz & 2yz & 2z^2 \end{bmatrix} P \qquad (3)$$

*3.2. SVM [13]*

A machine learning approach called a Support Vector Machine (SVM) is employed in classification and regression analysis. To keep sample points from various categories as distinct as possible, it classifies by locating the best feasible hyperplane in the feature space. It also exhibits high generalization for fresh, unlabeled sample points.

*3.3. Logistic regression [14]*

A popular statistical learning technique for creating categorization models is logistic regression. It predicts the binary classification issue and converts the data to a probability value between 0 and 1 by figuring out how a logistic function (such the Sigmoid function) and a linear combination of the input variables relate to one another.

*3.4. Fisher's Linear Discriminant [15]*

Finding the linear projection that best separates data from various classes may be done using the traditional pattern recognition technique known as linear discriminant analysis.

FLD aims to minimize the internal variance of samples from the same category while maximizing the separation of samples from other categories in the projected space. FLD can effectively classify objects in low-dimensional spaces in this way.

*3.5. Naive Bayes [16]*

Based on Bayes' theorem, Naive Bayes is a straightforward yet effective probabilistic classification technique. It makes the "naive" assumption that the traits are distinct from one another. When a particular feature vector is present, the Naive Bayes algorithm determines the posterior probability of each category and then chooses the category with the highest posterior probability as the prediction outcome.

*3.6. KNN [17]*

A straightforward and efficient classification algorithm is K-Nearest Neighbors. Based on the votes of its neighbors, it chooses which category the new sample points will be allocated to. The KNN algorithm's "K" stands for the number of nearest neighbors that have been chosen, and this number is derived from the distance metric.

*3.7. Loss function*

To examine the measurement of each model, four loss function models are introduced: the mean absolute error (MAE), mean square error (MSE), nonlinear loss measurement heteroscedastic effect positive MAE (HMAE), and tetra positive difference effect (HMSE).

$$MAE = \frac{1}{T}\sum |\hat{v}_t - RV_t|$$

$$MSE = \frac{1}{T}\sum (\hat{v}_t - RV_t)^2$$

$$HMAE = \frac{1}{T}\sum \left|1 - \frac{\hat{v}_t}{RV_t}\right|$$

$$HMSE = \frac{1}{T}\sum (1 - \frac{\hat{v}_t}{RV_t})^2$$

**4. Experiment**

*4.1. Selection of sample data*

To better compare the performance of the algorithm, this research uses random.rand in NumPy to randomly extract 1000 numbers, thereby generating random quaternion data and affixing two different labels.

## 4.2. Experimental procedure

The format of our research procedure is as follows. After generating a random dataset, convert the generated quaternion data into a rotation matrix and use it as an input feature. Among them, the labeled labels are binary categories and must be converted into numerical labels. Then the experiment will be divided into training set, verification set and test set with test_size=0.2 (drawing 20%), random_state=42 (control random state). In this study, we used five machine learning algorithms, namely SVM, logistic regression, FLD, naive Bayesian and KNN algorithms. After training the five models one by one, and combining the accumulated experience and debugging parameters, a certain prediction is made. Calculate the accuracy, recall and F1-score based on the predicted label, and calculate MSE, MAE, HMSE, HMAE, and draw a learning curve to observe whether the data is overfitting.

## 4.3. Result and Analysis

Through many observations on the learning curve, many adjustments were made to the experiment, and a relatively good record of experimental results was recorded. The experiment compared five models pairwise and recorded data using eight commonly used indicators of machine learning.

| Model | Precision | Recall | F1-score | Accuracy | MSE | MAE | HMAE | HMSE |
|---|---|---|---|---|---|---|---|---|
| Quaternion SVM | 0.554455 | 0.533333 | 0.543689 | 0.53 | 0.47 | 0.47 | 0.235 | 0.47 |
| Matrix SVM | 0.545455 | 0.4 | 0.461538 | 0.51 | 0.49 | 0.49 | 0.245 | 0.49 |
| Quaternion Logistic Regression | 0.553398 | 0.542857 | 0.548077 | 0.53 | 0.47 | 0.47 | 0.235 | 0.47 |
| Matrix Logistic Regression | 0.535354 | 0.504762 | 0.519608 | 0.51 | 0.49 | 0.49 | 0.245 | 0.49 |
| Quaternion FLD | 0.553398 | 0.542857 | 0.548077 | 0.53 | 0.47 | 0.47 | 0.235 | 0.47 |
| Matrix FLD | 0.55102 | 0.514286 | 0.53202 | 0.525 | 0.475 | 0.475 | 0.2375 | 0.475 |
| Quaternion Naive Bayes | 0.55 | 0.52381 | 0.536585 | 0.525 | 0.475 | 0.475 | 0.2375 | 0.475 |
| Matrix Naive Bayes | 0.528571 | 0.352381 | 0.422857 | 0.495 | 0.505 | 0.505 | 0.2525 | 0.505 |
| Quaternion KNN | 0.53125 | 0.485714 | 0.507463 | 0.505 | 0.495 | 0.495 | 0.2475 | 0.495 |
| Matrix KNN | 0.510204 | 0.47619 | 0.492611 | 0.485 | 0.515 | 0.515 | 0.2575 | 0.515 |

The different model indicators have essentially all somewhat improved from the usage of quaternions alone. When it comes to metrics like precision rate, F1 score, and recall rate, the two models of quaternion SVM and quaternion logistic regression perform about as well as they could. They have high precision rate and F1 score, and they can better detect true positive samples. These indications show a somewhat inferior performance for the two models, Matrix SVM and Matrix Logistic Regression, and low accuracy and F1 scores. Each model's accuracy falls within a narrow range, with little variation.

Quaternion SVM and Quaternion Logistic Regression may be preferable options if future researchers focus more on metrics like accuracy rate, F1 score, and recall rate, according to the study shown above. Even if minute variations may not matter as much in the grand scheme of things, they nonetheless have some bearing on model choice and application. Minor changes in normalization error may offer hints to model interpretability and uncertainty estimates, whereas minor differences in correctness may be significant in a specific position.

Therefore, it is still essential to carefully weigh these minor variations together with requirements, data set features, and application scenario considerations when choosing a model and planning.

## 5. Conclusion and discussion

According to the experimental findings, quaternions can help machine learning models perform better than conventional matrix-based techniques. This result is consistent with other studies that showed how useful quaternions are for a variety of tasks, especially when dealing with complicated spatial data and geometric transformations.

The better capacity of quaternion-based models to capture and describe spatial interactions may be a contributing factor to their higher performance. Quaternions are useful in fields like computer vision, robotics, and sensor fusion because they are adept at handling rotations and orientations. The models in this experiment were able to more effectively utilize the spatial information found in the dataset by including quaternion representations, which improved accuracy, precision, and F1 score.

It's crucial to remember, nevertheless, that not all datasets and workloads may benefit equally from the performance enhancements provided by quaternion-based models. The properties of the data and the needs of the current task have a significant impact on how well quaternions perform in machine learning. Before choosing to utilize quaternion-based models, researchers should carefully analyze the nature of their data and the domain in which they are working.

Future studies in this field may examine how well quaternion-based models perform over a larger range of datasets and tasks. Additionally, analyzing these models' interpretability and explain ability might offer insightful knowledge about how they function and contribute to the development of confidence in their predictions. To further improve the performance of quaternion-based models, efforts may be made to optimize the training methods and architectures created for such models.

The experiment concludes by highlighting the possible advantages of quaternions in machine learning models. Although in this study quaternion-based models outperformed matrix-based models in terms of performance, it is crucial to carefully consider the unique needs and peculiarities of the situation at hand when choosing and creating models. The results offer insightful information for the next study and applications, as well as to the ongoing investigation of quaternion-based machine learning systems.

**Acknowledgement**

This work is supported by Wenzhou Kean University Summer Student partnering with Faculty (SSPF) Research Program (WKUSSPF202312).